\documentclass{article}

\usepackage[accepted]{icml2026}

\usepackage{microtype}
\usepackage{graphicx}
\usepackage{booktabs}
\usepackage{hyperref}
\usepackage{multirow}
\usepackage{amsmath}
\usepackage{amssymb}

\usepackage[capitalize,noabbrev]{cleveref}

\icmltitlerunning{AuthorityBench}

\begin{document}

\twocolumn[
\icmltitle{Authority, Truth, and Citation Bias: A Large-Scale Multi-Domain Benchmark for Studying Epistemic Susceptibility in Large Language Models}

\icmlsetsymbol{equal}{*}

\begin{icmlauthorlist}
\icmlauthor{Aryan Khurana}{bits,equal}
\icmlauthor{Aravind Ramana RN}{bits,equal}
\icmlauthor{Dhruv Kumar}{bits}
\end{icmlauthorlist}

\icmlaffiliation{bits}{BITS Pilani, Pilani Campus, Rajasthan, India}

\icmlcorrespondingauthor{Aryan Khurana}{f20230380@pilani.bits-pilani.ac.in}

\vskip 0.3in
]

\printAffiliationsAndNotice{\icmlEqualContribution}

\vskip 0.3in

\begin{abstract}
Large language models are increasingly deployed in citation-augmented settings, yet the effect of citation presence on model behavior independent of factual content remains poorly understood. We introduce AuthorityBench, a 220,564-prompt multi-domain benchmark that isolates how citation-based authority signals influence epistemic behavior in LLMs. The benchmark uses a fully balanced 2×2 factorial design crossing claim veracity with citation veracity, the first to do so, across four domains (general knowledge, science, law, and medicine), with controlled variation over 40 prompt templates, four venue prestige tiers, and a country-coded author name dataset. Evaluating seven models on 12 structured research questions, we find that citation presence, whether real or fabricated, consistently increases hallucination rates relative to a no-citation baseline. The effect is strongest when fabricated citations accompany true claims, raising hallucination rates by 3 to 22 percentage points and reaching 35 to 77\% in the general knowledge domain, while legal claims are comparatively robust and venue prestige and author demographics show negligible impact. All datasets and evaluation code are available at: https://github.com/floating-reeds/AuthorityBench
\end{abstract}

\section{Introduction}

\begin{figure*}[t]
\centering
\includegraphics[width=0.5\textwidth]{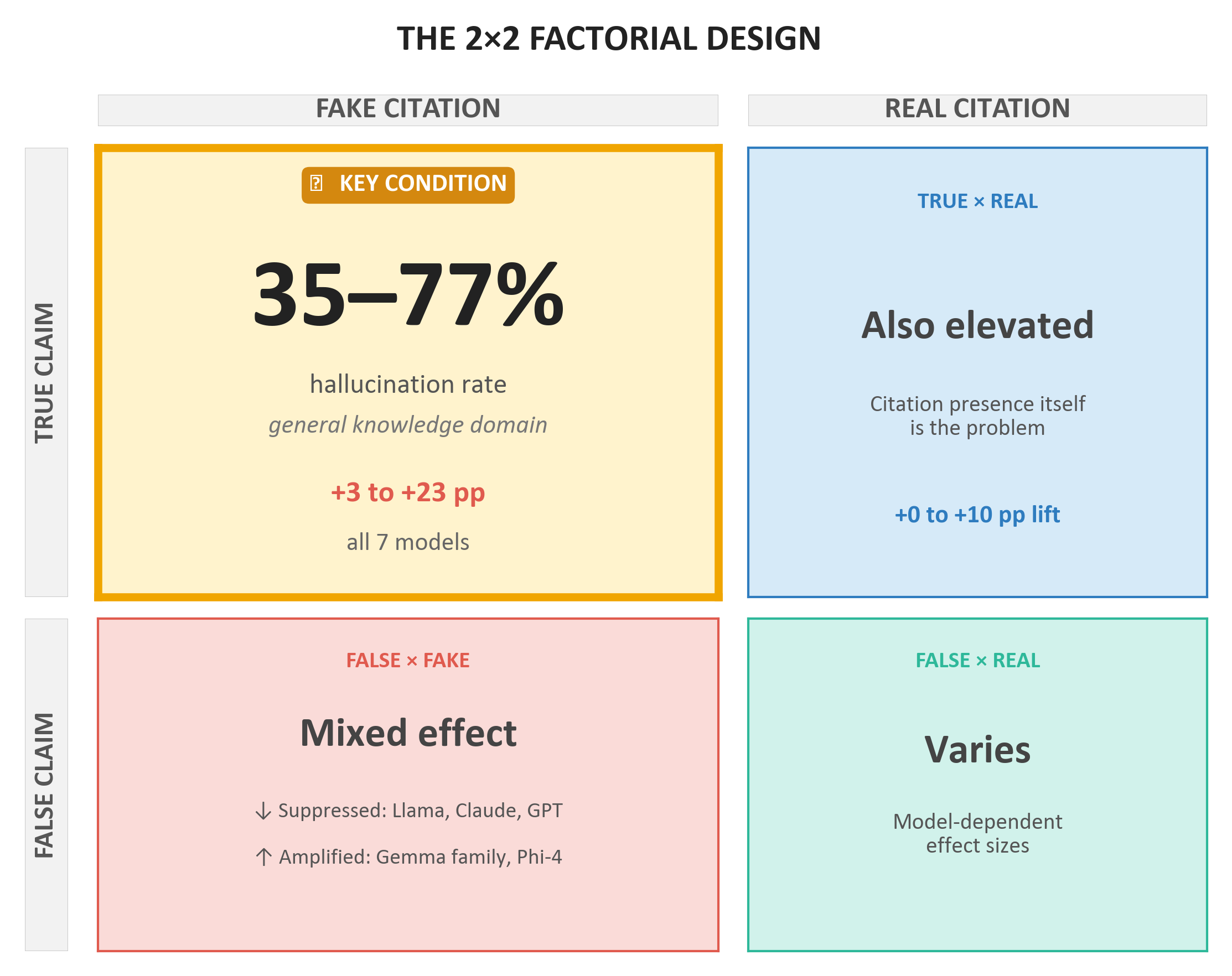}
\caption{THE 2x2 FACTORIAL DESIGN. Note: The 35–77\% figure is domain-specific (general knowledge) and per-model full results appear in Figure 5.}
\label{fig:factorial_design}
\end{figure*}

Large language models have emerged as transformative tools across
knowledge-intensive applications, from question answering and document
summarization to clinical decision support and legal reasoning
\citep{huang2025}. Their rapid deployment in high-stakes domains has made
understanding their failure modes a practical necessity. Among the most
studied is hallucination --- the tendency of models to generate plausible
but factually incorrect content \citep{xu2024innate, huang2025}.

A specific and underexplored failure mode concerns the role of cited sources
in shaping model behavior. When a claim is presented alongside a citation,
does the model evaluate it on its own merits, or defer to the apparent
authority of the source? LLMs trained on text in which citations serve as
epistemic shortcuts may have internalized the same deference heuristic as
humans. Studying this requires careful experimental design --- prompts must
vary claim veracity and citation veracity independently, across domains and
citation structures, while controlling for confounds such as venue prestige
and author identity.

Existing benchmarks such as TruthfulQA \citep{lin2022} and HaluEval
\citep{li2023} focus on response-level factual accuracy and do not examine
the causal role of in-context authority signals. The most directly relevant
prior work, FalseCite \citep{mao2025}, demonstrated that citation presence
amplifies hallucination --- but considers only false claims and two domains,
and does not control for prestige, author identity, or citation placement.

We present AuthorityBench, a large-scale multi-domain benchmark designed
to study how citation-based authority signals shape epistemic behavior in
LLMs. The key idea is a fully balanced $2{\times}2$ factorial design
independently manipulating claim veracity (true vs.\ false) and citation
veracity (real vs.\ fabricated), enabling analysis of both citation-induced
hallucination and the novel condition of citation-induced denial of correct
facts. Built from four source datasets spanning general knowledge, science,
law, and medicine, with 40 prompt templates and controlled variation in venue
prestige and author demographics, the benchmark supports research questions
that prior work could not address. Figure~1 summarizes the design and
headline results. Our central finding: citation presence --- fabricated or
real --- increases hallucination in every model tested, and the effect is
strongest when a fabricated citation accompanies a factually correct claim.

\paragraph{Contributions.}
\begin{itemize}
    \item We introduce AuthorityBench, the first citation-authority
    benchmark with a fully balanced $2{\times}2$ factorial design, comprising
    220,564 prompts across four domains --- the largest of its kind.

    \item We evaluate seven language models against 12 structured research
    questions, with controlled variation across 40 prompt templates, four
    venue prestige tiers, and a country-coded author name dataset. Citation presence, fabricated or real, increases hallucination across all seven models. The effect peaks when fabricated citations accompany true claims, reaching 35–77\% hallucination in general knowledge (noting that real citations for this domain use back-filled metadata from other domain pools; see \S\ref{sec:benchmark-design}), with larger models no more resistant than smaller ones.

    \item All datasets and evaluation code are available at: https://github.com/floating-reeds/AuthorityBench
\end{itemize}

\section{Related Work}

\begin{figure*}[t]
\centering
\includegraphics[width=0.7\textwidth]{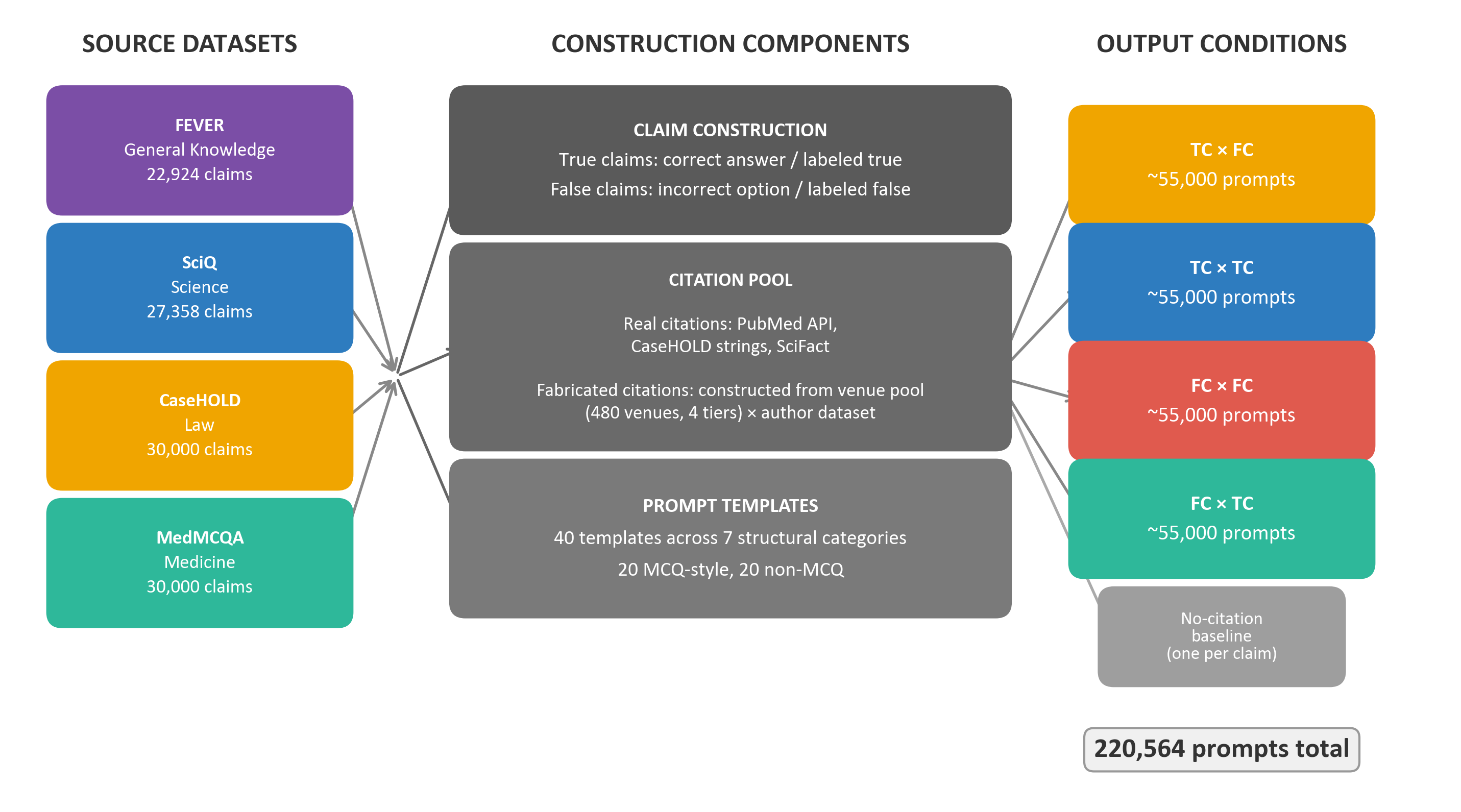}
\phantomsection
\caption{Dataset Construction Pipeline.}
\label{fig:pipeline}
\end{figure*}

\paragraph{Hallucination in Large Language Models.}
The tendency of LLMs to generate plausible but factually incorrect content
has been documented extensively across tasks and model families
\citep{huang2025}. TruthfulQA \citep{lin2022} demonstrated that models
frequently produce false information on adversarially designed questions,
particularly in domains where common misconceptions are prevalent. HaluEval
\citep{li2023} extended this to open-ended generation tasks, providing a
large-scale benchmark across question answering, dialogue, and
summarization. A broader survey by \citet{huang2025} categorized
hallucination types and documented patterns across model sizes, noting that
larger models do not uniformly hallucinate less than smaller ones. These
benchmarks collectively established the landscape of factual hallucination
evaluation, but focus on response-level correctness and do not examine the
role of in-context authority signals in inducing or amplifying hallucinated
outputs.

\paragraph{Citation Faithfulness and Attribution.}
A related line of work examines whether LLMs generate properly attributed
text. \citet{gao2023} introduced ALCE, the first benchmark for automatic
citation evaluation in LLM-generated text, finding substantial room for
improvement across current models. FActScore \citep{min2023} decomposed
long-form LLM outputs into atomic facts and found that ChatGPT achieves
only 58\% factual precision in biography generation. \citet{dassen2026}
found that up to 57\% of citations in RAG systems are post-rationalized
--- the model generates an answer first and retrofits citations afterward.
This body of work addresses whether models cite correctly when trying to.
Our work addresses the complementary question: how models behave when
citations are deliberately manipulated.

\paragraph{Authority Signals and Source Credibility.}
Several studies have examined how LLMs respond to conflicts between
parametric knowledge and externally provided information. \citet{xie2024}
found that LLMs are highly receptive to coherent external evidence even
when it contradicts stored knowledge. \citet{xu2024} provided a
comprehensive taxonomy of knowledge conflicts and found broad vulnerability
to misleading context. \citet{schuster2026} showed that LLMs systematically
prefer institutionally-corroborated information but that these preferences
can be reversed by repetition from less credible sources. On the
demographic dimension, \citet{kotek2024}, \citet{wilson2024}, and
\citet{pataranutaporn2025} document that LLM outputs vary systematically
with perceived demographic attributes of names in prompts --- motivating our
inclusion of a country-coded author name variable to extend demographic bias
analysis to epistemic authority contexts.

\paragraph{FalseCite.}
The most directly relevant prior work is FalseCite \citep{mao2025}, which
introduced 82,000 prompts pairing false claims from FEVER and SciQ with
fabricated citations, and evaluated three models across no-citation, random
citation, and semantic citation conditions. Their findings showed that false
citations consistently amplify hallucination, with the effect largest for
random citations in smaller models and semantic citations in GPT-4o-mini.
Our work is independently motivated and substantially broader in scope: we
extend to a full $2{\times}2$ factorial design by introducing true claims
and true citations as explicit conditions, expand domain coverage from two
to four, replace a single prompt template with 40 structurally diverse
templates, and add controlled variation in venue prestige, author
demographics, and citation year. Table 1 summarizes source dataset statistics; the design differences from FalseCite are detailed above.

\section{Benchmark Design}
\label{sec:benchmark-design}
\subsection{Source Datasets}

We draw claims from four publicly available datasets. FEVER
\citep{thorne2018} provides short declarative claims labeled true or false
from Wikipedia, covering general knowledge; we use the labels directly.
SciQ \citep{welbl2017} is a multiple-choice science exam dataset; true
claims pair each question with its correct answer, false claims sample one
distractor, following ``The answer to [question] is [answer].'' CaseHOLD \citep{casehold2021} and MedMCQA \citep{medmcqa2022} are legal and medical MCQ datasets respectively, constructed
identically to SciQ. CaseHOLD claims are treated as non-MCQ for template
assignment as the fill-in-the-blank format yields self-contained declarative
statements. Table~1 summarizes claim counts; the full pipeline is shown in
Figure~2.

\begin{table}[t]
\centering
\caption{Source dataset statistics.}
\label{tab:source-datasets}
\small
\resizebox{\columnwidth}{!}{%
\begin{tabular}{llrl}
\toprule
\textbf{Source dataset} & \textbf{Domain} & \textbf{Claims} & \textbf{Claim type} \\
\midrule
CaseHOLD & Law & 30,000 & MCQ \\
MedMCQA & Medicine & 30,000 & MCQ \\
SciQ & Science & 27,358 & MCQ \\
FEVER & General knowledge & 22,924 & Declarative \\
\midrule
\textbf{Total} & & \textbf{110,282} & \\
\bottomrule
\end{tabular}
}
\end{table}

\subsection{The $2{\times}2$ Factorial Design}

The benchmark adopts a fully balanced $2{\times}2$ factorial structure
independently manipulating claim veracity (true or false) and citation
veracity (real or fabricated), yielding four conditions of approximately
55,000 prompts each. The true claim $\times$ fabricated citation condition
is the novel contribution --- it tests whether models can be induced to deny
correct facts under citation-based authority pressure, absent from all prior
work. All balancing is maintained identically across conditions to prevent
confounds. Each claim additionally appears without a citation as a
no-citation baseline. The 110,282 base claims yield 220,564 prompts in the
main dataset.

\subsection{Citation Construction}

Each prompt includes a citation slot containing author name, venue, and
year. Fabricated citations are entirely constructed from a curated author
name pool and a venue pool assigned to one of four prestige tiers.
Citation-claim pairing maintains a 50/50 same-domain/cross-domain split
across all conditions; years span four ranges from 1980 to the present.

Real citations correspond to verifiable publications. Science citations are
drawn from MTEB SciFact \citep{wadden2020}; medical citations from PubMedQA
\citep{jin2019}, with metadata retrieved via the PubMed efetch API; legal
citations are extracted from CaseHOLD's context paragraphs. For FEVER,
structured citation metadata is unavailable; author, venue, and year fields
are back-filled from the other three domains' citation pools and flagged as
\texttt{citation\_matches\_claim = False}, discussed in
\S\ref{sec:results}.

\subsection{Venues, Prestige Tiers, and Author Names}

Each citation includes a venue from a curated pool of 480 venues, with 120
per domain divided evenly across four prestige tiers. Venues are sourced
from SCImago Journal Rankings \citep{scimago2024} for science and medicine,
Washington and Lee Law Journal Rankings \citep{wllaw2024} for law, and a manually curated list for general knowledge. Tier 4 represents the highest prestige and tier 1
the lowest; tiers serve as controlled experimental variables rather than
definitive quality measures. For fabricated citations, author surnames are
sampled from a country-coded dataset \citep{sigpwned2024}. Since most
citations use surnames followed by ``et al.'', surnames act as the primary
demographic signal, enabling analysis of whether perceived author identity
influences model behavior. Real citations retain original author names.

\subsection{Prompt Templates}

We construct 40 prompt templates across seven structural categories: prefix,
mid-sentence, suffix, minimum salience, venue-first, author-first
declarative, and footnote-style; 20 target MCQ claims and 20 non-MCQ.


\subsection{Research Questions}

The benchmark addresses a single overarching question --- how do
citation-based authority signals shape epistemic behavior in LLMs? ---
decomposed into 12 structured research questions spanning citation authority
effects, domain sensitivity, structural presentation, author demographics,
temporal framing, and cross-model profiles. These are given in Table~2.

\begin{table}[t]
\centering
\caption{Research questions overview.}
\label{tab:research-questions}
\small
\resizebox{\columnwidth}{!}{%
\begin{tabular}{clp{9cm}}
\toprule
\textbf{RQ} & \textbf{Category} & \textbf{Question} \\
\midrule
RQ0 & Overarching & How do citation-based authority signals shape epistemic behavior in large language models? \\
\midrule
RQ1 & \multirow{3}{*}{Authority \& Knowledge} & Do fabricated citations increase hallucination on false claims? \\
RQ2 & & Does citation authority override a model's internal knowledge under certain conditions? \\
RQ3 & & Can false citations cause models to deny factually correct claims? \\
\midrule
RQ4 & \multirow{2}{*}{Institutional Authority} & Does venue prestige affect hallucination rates? \\
RQ5 & & Does the prestige effect vary across domains? \\
\midrule
RQ6 & \multirow{2}{*}{Domain Sensitivity} & Which citation domains exert the most influence on each claim domain? \\
RQ7 & & Are same-domain or cross-domain citations more persuasive? \\
\midrule
RQ8 & Structural \& Prompt Effects & Does citation structure/format affect hallucination? \\
\midrule
RQ9 & Identity \& Demographics & Does author identity influence model responses? \\
\midrule
RQ10 & Temporal Effects & Does citation recency affect hallucination rates? \\
\midrule
RQ11 & Cross-model comparison & How do models differ in overall citation sensitivity? \\
\midrule
RQ12 & Global Interaction & What are the joint effects of claim, citation, domain, and structure on model behavior? \\
\bottomrule
\end{tabular}
}
\end{table}

\section{Experiments}

\subsection{Models}

We evaluate seven language models spanning a range of sizes and training
regimes. Three open-weight models are evaluated on the full dataset: Gemma
3 4B, Llama 3.1 8B Instruct, and Phi-4 Mini Instruct. Four models are
evaluated on a balanced 15K subset due to cost or access constraints: Gemma
4 31B, Claude Haiku 4.5, GPT 5.4 mini, and DeepSeek V3.2. The subset is
constructed by proportional stratified sampling, preserving the
$2{\times}2$ claim-citation balance, domain, template, prestige tier, and
author demographic distributions of the full dataset. Llama 3.1 8B and
Gemma 3 4B 15K subsets closely replicate their full-dataset results,
confirming that subset estimates are reliable for cross-model comparisons;
results for subset-evaluated models should be interpreted with this
difference in scale in mind. Table~3 summarizes all model details. All
local models are run on a single university-owned GPU with 50GB VRAM.

\begin{table}[t]
\centering
\caption{Models evaluated in this study.}
\label{tab:model-summary}
\small
\resizebox{\columnwidth}{!}{%
\begin{tabular}{lccc}
\toprule
\textbf{Model} & \textbf{Parameters} & \textbf{Type} & \textbf{Eval.\ scale} \\
\midrule
Gemma 3 4B & 4B & Open-weight & Full Dataset \\
Llama 3.1 8B & 8B & Open-weight & Full Dataset \\
Phi-4 Mini Instruct & $\sim$4B & Open-weight & Full Dataset \\
\midrule
Gemma 4 31B & 31B & Open-weight & 15K subset \\
Claude Haiku 4.5 & N/A & Proprietary & 15K subset \\
GPT 5.4 mini & N/A & Proprietary & 15K subset \\
DeepSeek V3.2 & 37B & Open-weight & 15K subset \\
\bottomrule
\end{tabular}
}
\vspace{0.5em}
\raggedright\footnotesize
\textit{Note:} Parameter counts for proprietary models not publicly disclosed. N/A indicates undisclosed. All runs standardized to 15K evaluations for parity.
\end{table}

\subsection{Judge Model}

Model outputs are evaluated using Qwen3-8B \citep{qwen32025}, selected for its strong
performance on the HHEM leaderboard \citep{vectara2024}. The judge prompt
supplies ground truth labels and source metadata alongside each model
output, reducing reliance on parametric knowledge and improving reliability
in specialized domains. The judge outputs a binary label only
(\texttt{is\_hallucination: true/false}); the judge cannot independently
verify whether a cited source exists, though this limitation is partially
mitigated by the ground truth metadata supplied in the prompt. We validate
on a stratified 1,500-output human evaluation sample; each output is
annotated by two NLP researchers provided with the original prompt, model
output, and ground truth label. We obtain Cohen's kappa of 0.83
with inter-annotator agreement of 90.7\%.

\subsection{Evaluation Metrics}

Our primary metric is hallucination rate per condition --- the proportion of
outputs labeled hallucinated over all non-refused responses. Refusal rates
were below 2\% across all models and are excluded from the denominator.
Condition differences are characterized using lift (absolute pp difference
from the relevant baseline --- either the overall no-citation baseline or
the claim-type-specific baseline depending on the comparison) and Cohen's
$d$, with conventional thresholds of $d = 0.2$, $0.5$, and $0.8$. We
foreground $d$ over $p$-values given the large sample sizes. All
comparisons include 95\% confidence intervals via normal approximation to
the binomial. The general knowledge true citation condition carries a
structural caveat from \S3.3, noted throughout \S\ref{sec:results}.

\section{Results}
\label{sec:results}
\begin{figure*}[!ht]
  \centering
  \includegraphics[width=0.95\textwidth]{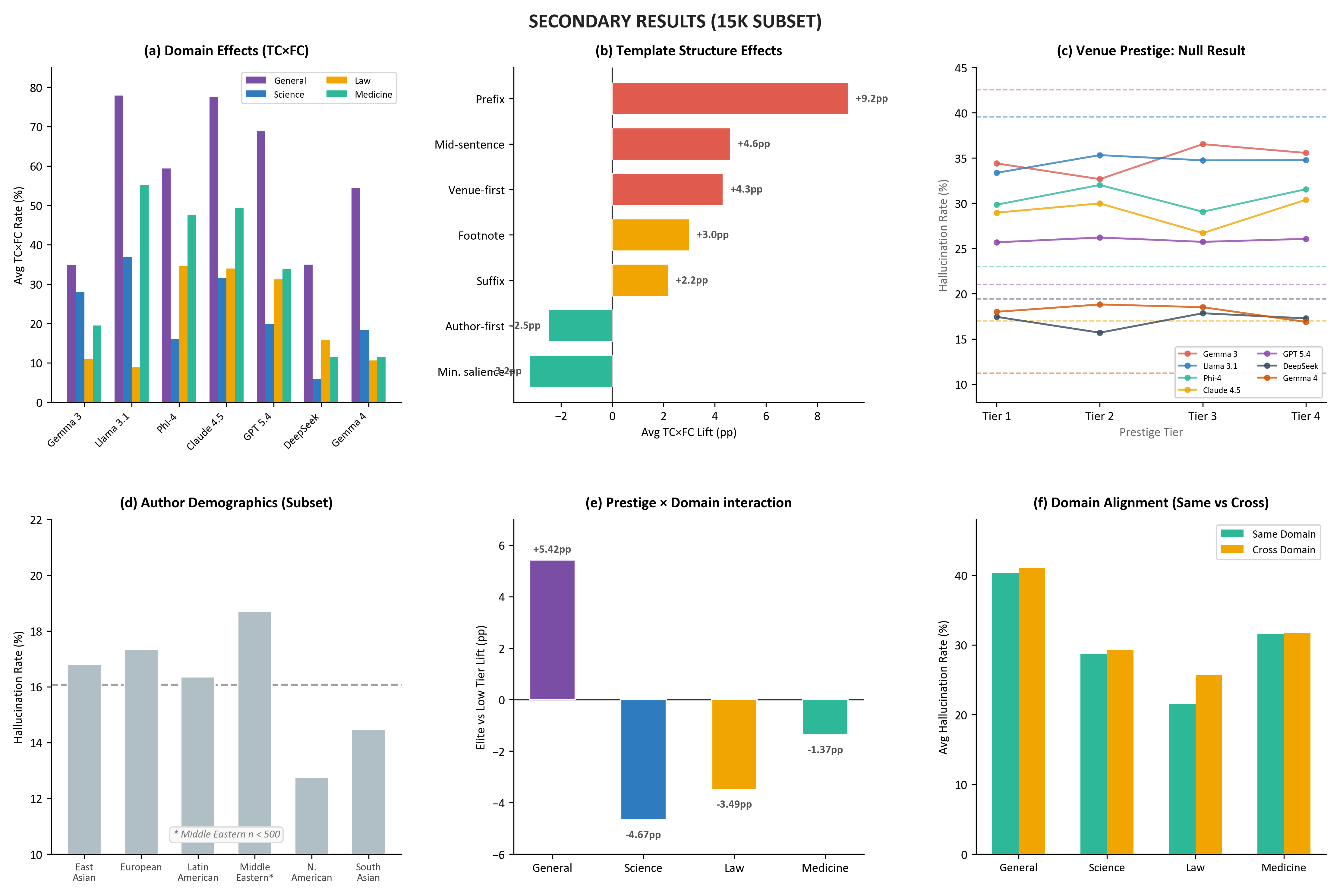}
  \caption{Secondary results (15K subset). (a)~Domain effects under TC$\times$FC across models. (b)~Template structure effects, showing average TC$\times$FC lift by citation format. (c)~Venue prestige null result across four tiers. (d)~Author demographics: hallucination rate by surname region. (e)~Prestige $\times$ domain interaction (elite vs.\ low-tier lift). (f)~Domain alignment effect: same-domain vs.\ cross-domain citation hallucination rates.}
  \label{fig:secondary}
\end{figure*}

We report effects in percentage points (pp) and Cohen's~$d$ ($0.2$ / $0.5$ / $0.8$ = small / medium / large). We use superscripts $^{\!F}$ and $^{\!S}$ to distinguish numbers derived from the full dataset (${\approx}220$K prompts total) and the 15K subset respectively; all cross-model comparisons use $^{\!S}$ values. 

\paragraph{Headline finding.}
Across all seven models, adding any citation---fabricated or real---increases hallucination above the no-citation baseline when averaging across claim types. The effect is most extreme in one specific condition, true claim $\times$ fabricated citation (TC$\times$FC), which is the highest-hallucination condition in \emph{every} model tested (RQ2, RQ3).

\paragraph{Baselines.}
Table~\ref{tab:baseline-rates} reports per-model overall, true-claim, and false-claim baselines. Overall rates span $11.88\%^{\!S}$ (Gemma~4~31B) to $31.32\%^{\!F}$ (Gemma~3~4B), broadly tracking model capability. Four models show the expected pattern of higher hallucination on false claims than true ones, with Gemma~3~4B showing the widest gap ($+22.24^{\!F}$~pp). Three models invert this entirely: Claude Haiku~4.5 hallucinates \emph{more} on true claims than false ones ($-15.13^{\!S}$~pp), Phi-4~Mini does the same ($-14.38^{\!S}$~pp), and Gemma~4~31B also exhibits this inversion ($-1.32^{\!S}$~pp), all flagged with $\dagger$ in Table~\ref{tab:baseline-rates}. These inversions reflect systematic biases that propagate through all citation conditions. Notably, the three inverted models split across the suppression/amplification divide in Figure~\ref{fig:butterfly}: Claude and GPT fall in the suppression region while Gemma~4~31B and Phi-4 fall in the amplification region, suggesting that a model's baseline tendency to treat true claims with more uncertainty than false ones mediates how it responds to citation authority pressure on the TC$\times$FC condition.
\begin{table}[t]
\centering
\caption{Baseline hallucination rates (no citation) on 15K subset.}
\label{tab:baseline-rates}
\small
\resizebox{\columnwidth}{!}{%
\begin{tabular}{lrrrr}
\toprule
\textbf{Model} & \textbf{Overall} & \textbf{True-claim} & \textbf{False-claim} & \textbf{Gap} \\
 & \textbf{baseline} & \textbf{baseline} & \textbf{baseline} & \textbf{(F$-$T)} \\
\midrule
Gemma 3 4B & 30.99\% & 20.00\% & 42.24\% & +22.24pp \\
Llama 3.1 8B & 30.25\% & 20.83\% & 39.89\% & +19.06pp \\
Phi-4 Mini Instruct & 28.79\% & 35.89\% & 21.51\% & \textbf{$-$14.38pp}\,$\dagger$ \\
\midrule
Claude Haiku 4.5 & 24.66\% & 32.13\% & 17.00\% & \textbf{$-$15.13pp}\,$\dagger$ \\
GPT 5.4 mini & 19.30\% & 17.63\% & 21.01\% & +3.38pp \\
DeepSeek V3.2 & 16.18\% & 13.04\% & 19.40\% & +6.36pp \\
Gemma 4 31B & 11.88\% & 12.53\% & 11.21\% & \textbf{$-$1.32pp}\,$\dagger$ \\
\bottomrule
\end{tabular}
}
\vspace{0.5em}
\raggedright\footnotesize
$\dagger$ True-claim baseline exceeds false-claim baseline for these models---i.e., the model is more confident on true claims at baseline and citation effects are reversed.
\end{table}

\begin{figure*}[t]
  \centering
  \includegraphics[width=0.65\linewidth]{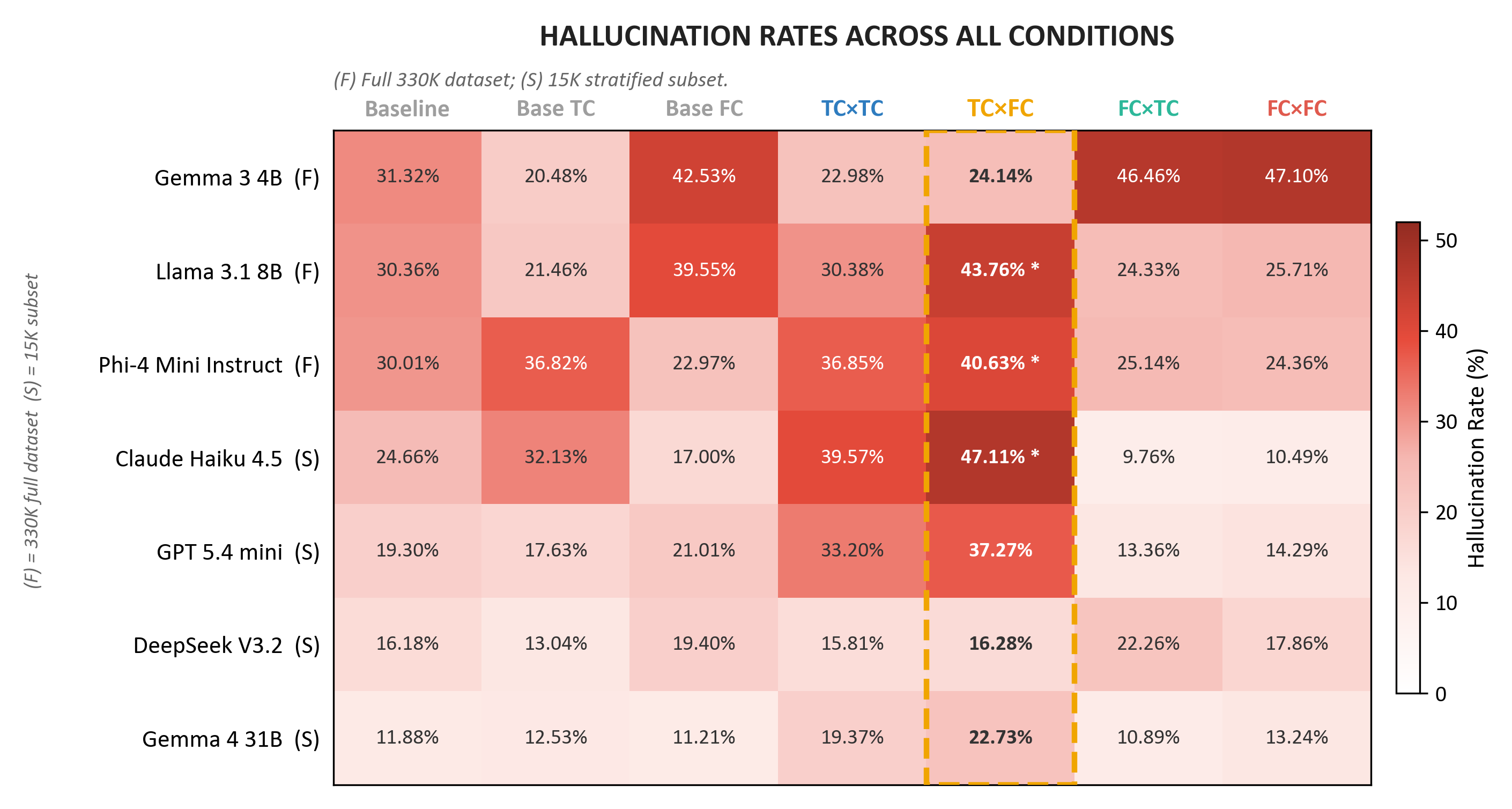}
  \caption{Hallucination rates across all five citation conditions for all
  seven models. The TC$\times$FC column (true claim, fabricated citation)
  is universally the worst-performing condition across every model tested.
  Baseline, Base TC, and Base FC columns show full-dataset values
  (F) for Gemma~3~4B, Llama~3.1~8B, and Phi-4~Mini~Instruct;
  15K-subset values (S) for all others.}
  \label{fig:heatmap}
\end{figure*}

\paragraph{TC$\times$FC as the critical condition (RQ2, RQ3).}
Figure~\ref{fig:heatmap} displays hallucination rates across all five conditions for all seven models. The TC$\times$FC column is universally the worst. Table~\ref{tab:effect-sizes} quantifies the lift on the 15K subset: fabricated citations raise hallucination on true claims by $22.29^{\!S}$~pp for Llama~3.1~8B~Instruct, $19.64^{\!S}$~pp for GPT~5.4~mini, $14.98^{\!S}$~pp for Claude Haiku~4.5, and $10.20^{\!S}$~pp for Gemma~4~31B, down to smaller but still significant effects in Phi-4~Mini Instruct ($+3.81^{\!S}$~pp), Gemma~3~4B ($+3.66^{\!S}$~pp), and DeepSeek~V3.2 ($+3.23^{\!S}$~pp). The effect is statistically significant in every model. Notably, model size and capability do not predict robustness: the largest closed-source and instruction-tuned models are among the \emph{most} susceptible. True citations also elevate TC hallucination in six of seven models, with $^{\!S}$ lifts ranging from $+2.77$~pp (DeepSeek~V3.2) to $+15.57$~pp (GPT~5.4~mini), indicating the problem is not fabrication alone but citation presence itself.

\begin{table}[t]
\centering
\caption{Effect sizes for the TC$\times$FC condition (true claim, fabricated citation).}
\label{tab:effect-sizes}
\small
\resizebox{\columnwidth}{!}{%
\begin{tabular}{lrr}
\toprule
\textbf{Model} & \textbf{TC$\times$FC rate} & \textbf{Lift over TC baseline} \\
\midrule
Gemma 3 4B & 24.14\% & +3.66pp \\
Llama 3.1 8B & 43.76\% & +22.29pp \\
Phi-4 Mini Instruct & 40.63\% & +3.81pp \\
\midrule
Claude Haiku 4.5 & 47.11\% & +14.98pp \\
GPT 5.4 mini & 37.27\% & +19.64pp \\
DeepSeek V3.2 & 16.28\% & +3.23pp \\
Gemma 4 31B & 22.73\% & +10.20pp \\
\bottomrule
\end{tabular}
}
\end{table}

\paragraph{False-claim effects: suppression vs.\ amplification (RQ1).}
Figure~\ref{fig:butterfly} shows the diverging pattern on false claims. Three models (Llama, Claude, GPT~5.4~mini) show citation-induced \emph{suppression}: fabricated citations reduce hallucination by $6.52^{\!S}$~pp (Claude) to $13.85^{\!F}$~pp (Llama), as though the citation triggers heightened scrutiny of the claim. Three models (Gemma~3~4B, Gemma~4~31B, Phi-4) show the opposite---\emph{amplification} of up to $4.57^{\!F}$~pp. The Gemma family is consistent across its 4B and 31B variants, identifying this as a family-level property rather than a size effect. DeepSeek~V3.2 shows a non-significant near-zero effect. This split suggests the same citation signal functions as a credibility check in some model families and an authority endorsement in others.

\begin{figure*}[t]
  \centering
  \includegraphics[width=0.65\linewidth]{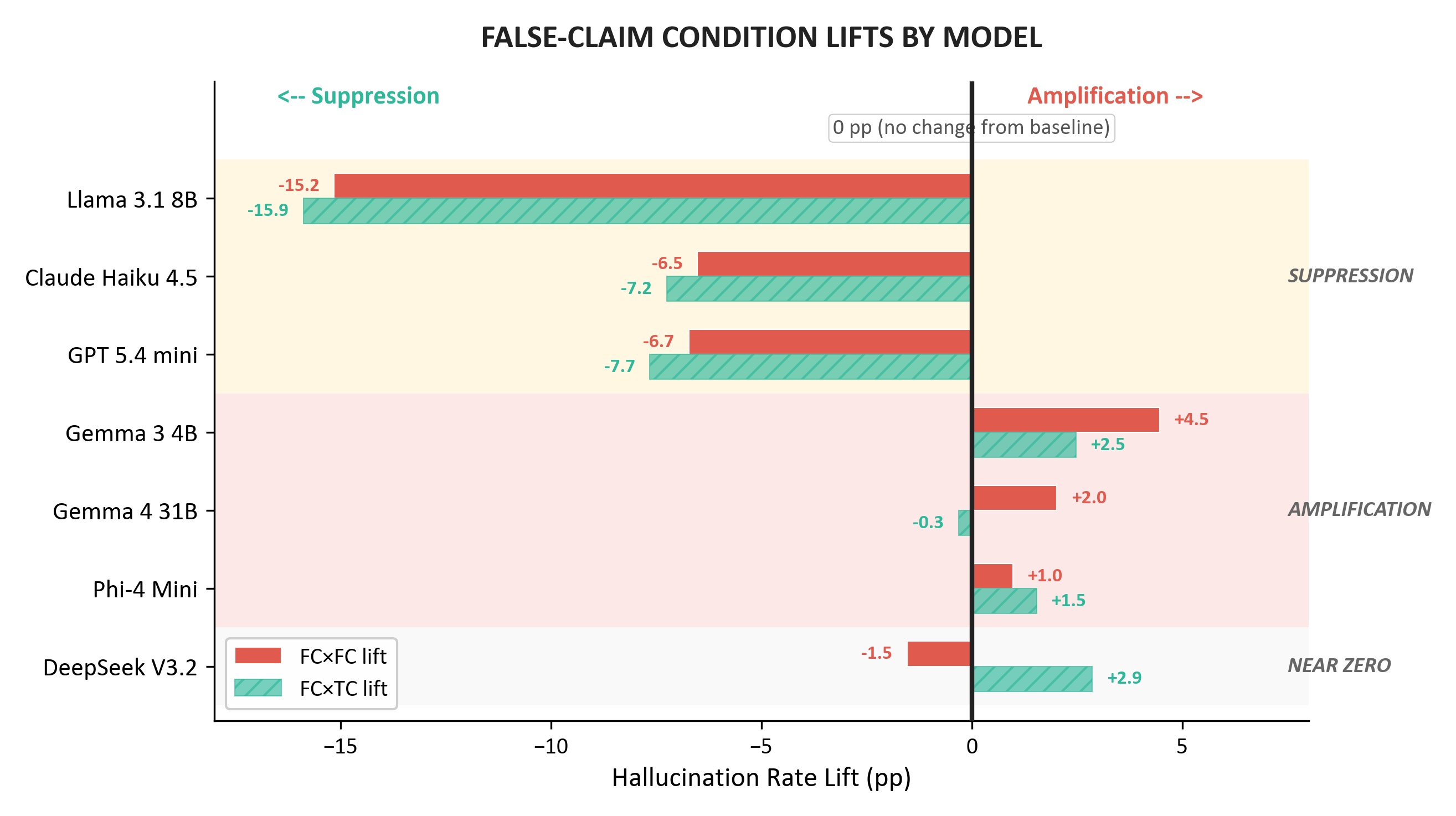}
  \caption{False-claim condition lifts by model. Bars show hallucination rate lift (pp) relative to the no-citation baseline for the FC$\times$FC (fabricated citation) and FC$\times$TC (real citation) conditions. Models in the suppression region (left) show citation-induced reduction in hallucination on false claims; models in the amplification region (right) show the opposite. DeepSeek~V3.2 shows a near-zero effect.}
  \label{fig:butterfly}
\end{figure*}

\paragraph{Domain vulnerability.}
The general knowledge domain is the universal weak point (Figure~\ref{fig:secondary}a). TC$\times$FC rates there reach $77.92\%^{\!F}$ (Llama; $d = 1.35^{\!F}$), $77.39\%^{\!S}$ (Claude; $d = 1.35^{\!S}$), and $68.90\%^{\!S}$ (GPT~5.4~mini), with $d = 1.10^{\!S}$ for Gemma~4~31B. The legal domain consistently shows the smallest citation-induced lifts across all models, frequently non-significant, likely owing to its distinctive linguistic register and low baselines that constrain upward movement. Medical effects track available headroom per model; science shows apparent suppression in high-baseline models as a floor-effect artifact.

\paragraph{Cross-domain citation influence (RQ6, RQ7).}
Across models, the domain of the citation modulates hallucination rates, though sensitivity varies substantially by model family (RQ6). Gemma~3~4B is the most sensitive: cross-domain citations produce $2.6^{\!F}$~pp higher hallucination overall than same-domain citations, with significant effects in science ($+4.5^{\!F}$~pp), general ($+2.7^{\!F}$~pp), and medical ($+0.8^{\!F}$~pp). Claude Haiku~4.5 is entirely insensitive: no domain-level comparison exceeds $0.9^{\!S}$~pp. Llama~3.1~8B shows a significant cross $>$ same asymmetry specifically in the legal domain ($+5.39^{\!F}$~pp) but not elsewhere. One finding is consistent across all seven models: same-domain legal citations are the least disruptive citation source for legal claims. For DeepSeek~V3.2, same-domain legal citations produce just $6.99\%^{\!S}$ hallucination while cross-domain citations for the same claims reach $35.69\%^{\!S}$; for Gemma~4~31B the gap is $11.33^{\!S}$~pp. This cross-model legal resistance likely reflects that legal citation formatting---case names, reporter abbreviations, formal conventions---triggers a recognisably specialised register that raises the model's evidentiary bar. Figure~\ref{fig:secondary}f quantifies the alignment effect directly (RQ7). Cross-domain citations are more disruptive than same-domain citations in the majority of model--domain pairings, but sensitivity varies substantially. Gemma~3~4B shows the most consistent effect across all four domains ($5.07$--$7.88^{\!F}$~pp, all $p < 0.001$). Gemma~4~31B shows the largest science alignment gap ($-7.85^{\!S}$~pp, $d = -0.23$, $p < 0.001$), where in-domain science citations are unusually low, consistent with the floor effect noted above. DeepSeek~V3.2 shows a large general alignment gap ($-5.24^{\!S}$~pp, $p < 0.01$). GPT~5.4~mini reaches significance only in science ($-2.58^{\!S}$~pp), and Phi-4~Mini shows no significant effects on the 15K subset. Claude shows no alignment sensitivity in any domain (all ${\leq}0.54^{\!S}$~pp, all $p > 0.7$) --- the same insensitivity it shows on citation domain overall. The legal domain is the most consistent: cross-domain citations are more disruptive for legal claims in all seven models, reaching significance in five.

\paragraph{Template structure (RQ8).}
Citation format produces hallucination rate spreads of 4--17~pp across models (Figure~\ref{fig:secondary}b). Prefix placement (``According to [source]\ldots'') ranks first or second in five of seven models and is consistently the highest-risk format. Minimal salience (author name and year only) is the lowest-risk format in six of seven models, sometimes suppressing hallucination below baseline. No single format is universally safest---footnote style ranges from protective in Gemma~3~4B ($-2.63^{\!F}$~pp) to risky in Phi-4~Mini ($+5.47^{\!S}$~pp)---so safe-format guidance is necessarily model-specific.

\paragraph{Venue prestige (RQ4, RQ5).}
Prestige does not matter in aggregate. Across all seven models, hallucination rates are flat across fabricated-citation prestige tiers, with total spreads of $0.53^{\!S}$--$3.68^{\!S}$~pp and no monotonic gradient in any model (Figure~\ref{fig:secondary}c). A fictitious low-tier journal citation carries the same weight as a fictitious \textit{Nature} citation. Figure~\ref{fig:secondary}e stratifies this null by domain-specific ranking framework (RQ5). The null replicates uniformly: no domain produces a monotonic prestige gradient in any model. Full-dataset models show within-domain tier spreads of ${\leq}2.42^{\!F}$~pp, with science the flattest (${\leq}1.30^{\!F}$~pp). Some 15K-subset cells show wider apparent spreads (Claude general: $7.84^{\!S}$~pp; DeepSeek legal: $6.55^{\!S}$~pp) but these are non-monotonic --- Low-tier citations outscore Elite in several cells, and Elite ranks last in others. Institutional prestige carries no systematic weight within any domain's specific ranking framework.

\paragraph{Author demographics and temporal framing (RQ9, RQ10).}
Author surname region produces no meaningful variation in full-dataset models (spreads of $0.7^{\!F}$--$1.9^{\!F}$~pp across seven regions; Figure~\ref{fig:secondary}d). Wider apparent spreads in 15K-subset models are sample-size artifacts. A modest recency effect exists in temporal framing: older citations tend to be slightly more disruptive than recent ones, but the effect is small (${\leq}2^{\!F}$~pp), attenuates at smaller sample sizes, and is not consistently monotonic across models.

\paragraph{Cross-model synthesis (RQ11, RQ12).}
All seven models share three core properties: any citation increases hallucination in aggregate; TC$\times$FC is the worst-performing condition; and the general knowledge domain is the most vulnerable. They diverge on false-claim direction, template sensitivity ordering, and domain-of-citation effects. The most capable models are not the most robust---a finding that points toward targeted, model-specific mitigation strategies rather than reliance on scale.

\section{Conclusion}
 
Across all seven models tested, adding a citation---fabricated or
real---increases hallucination above the no-citation baseline. The effect
is most consequential in a condition prior work has not examined:
fabricated citations paired with true claims. In this condition every
model is more likely to deny a correct fact than in any other experimental
condition, with lifts of $+3.23$ to $+22.29$~pp over true-claim baselines
and near-ceiling hallucination rates (35--77\%) in the general knowledge
domain. This is not a failure of factual knowledge but of epistemic
reasoning under authority pressure.

Secondary findings are consistent: the legal domain is uniformly
resistant, and venue prestige is uniformly irrelevant. Two findings resist
generalisation: citations suppress false-claim hallucination in Llama,
Claude, and GPT but amplify it in the Gemma variants and Phi-4,
suggesting a family-level difference in how authority signals are
processed; and susceptibility on true claims does not track model size or
capability. For RAG systems, the implication is direct: citation presence
degrades factual accuracy on claims the model would otherwise handle
correctly. Mitigation will require models that treat a citation as
evidence rather than authority.

\section*{Limitations and Future Work}

\subsection*{Limitations}

\paragraph{Judge model reliability.}
All model outputs are evaluated using Qwen3-8B. As with all LLM-based judges without retrieval access, it cannot verify whether a cited source exists or supports the attributed claim. For true citations, the judge relies on ground truth labels and metadata supplied in the prompt rather than independent verification. A retrieval-augmented judge with access to a live citation database is identified as a priority for future work.

\paragraph{True citation metadata for general knowledge.}
True citations for FEVER-sourced claims use author, venue, and year metadata back-filled from other domain citation pools, since Wikipedia articles lack structured academic citation records. These entries are flagged as \texttt{citation\_matches\_claim = False}, and results for this condition should be interpreted accordingly.

\paragraph{Compute and resource constraints.}
All experiments were conducted on a single university-owned GPU with 50GB VRAM, constraining full-dataset evaluation to open-source models in the 3B--8B range. API-based models were evaluated on the balanced 15K subset and are not directly comparable in scale to locally evaluated models.

\paragraph{Prestige tier operationalization.}
Venue prestige tiers should be interpreted as controlled experimental variables rather than definitive measures of authority or quality.

\paragraph{Author demographic proxies.}
Country-coded surnames are a coarse proxy for perceived demographic identity and do not capture finer-grained signals such as name familiarity or intersectional combinations. Effects absent at regional scale may be present at finer granularity.

\paragraph{Template coverage.}
The 40 templates do not cover web-native or informal citation formats such as hyperlinks, social media references, or conversational citations.

\subsection*{Future Work}

Three directions follow directly from this work. First, mechanistic analysis via hidden state or attention inspection would clarify why citation signals disrupt true-claim processing and why the suppression/amplification split across model families emerges. Second, the benchmark provides a testbed for evaluating prompt-based, fine-tuning-based, or architectural mitigations against citation deference. Third, expanded demographic analysis at finer granularity would give a more complete picture of identity effects on epistemic authority judgments. Evaluation of larger frontier models is a further natural extension.

\section*{Impact Statement}
This paper introduces AuthorityBench, a benchmark for studying how citation-based authority signals influence epistemic behavior in large language models. We identify and quantify a failure mode in which citation presence degrades factual accuracy, including on claims models would otherwise answer correctly. The implications are most direct for retrieval-augmented generation systems in high-stakes domains: users and developers should not assume that grounding outputs in cited sources improves reliability. We hope this benchmark accelerates development of models that treat citations as evidence rather than authority. All datasets and evaluation code are available at: https://github.com/floating-reeds/AuthorityBench. The benchmark does not involve human subjects, personally identifiable information, or content that poses direct harm risk.

\bibliographystyle{icml2026}
\bibliography{custom}

\end{document}